%% file: main.tex
\documentclass{article}

\PassOptionsToPackage{numbers, compress}{natbib}

\usepackage[preprint]{neurips_2023}

\usepackage[utf8]{inputenc} 
\usepackage[T1]{fontenc}    
\usepackage{hyperref}       
\usepackage{url}            
\usepackage{booktabs}       
\usepackage{amsfonts}       
\usepackage{nicefrac}       
\usepackage{microtype}      
\usepackage{xcolor}         
\usepackage{xspace}
\usepackage{mdframed}
\usepackage{wrapfig}
\usepackage{subfigure}
\usepackage{graphicx}

\input{macros}

\title{SLiC-HF: Sequence Likelihood Calibration with Human Feedback}

\author{%
  Yao Zhao$^\dagger$ \\ \texttt{yaozhaoyz@google.com} \And
  Rishabh Joshi$^\dagger$ \\ \texttt{rishabhjoshi@google.com} \And
  Tianqi Liu$^*$ \\ \texttt{tianqiliu@google.com} \AND
  Misha Khalman$^\dagger$ \\ \texttt{khalman@google.com} \And
  Mohammad Saleh$^\dagger$ \\ \texttt{msaleh@google.com} \And
  Peter J. Liu$^\dagger$ \\ \texttt{peterjliu@google.com} \AND
  \normalfont{Google Deepmind$^\dagger$, Google Research$^*$}
}

\begin{document}

\maketitle

\begin{abstract}
Learning from human feedback has been shown to be effective at aligning language models with human preferences.
Past work has often relied on Reinforcement Learning from Human Feedback (RLHF), which optimizes the language model using reward scores assigned from a reward model trained on human preference data.
In this work we show how the recently introduced Sequence Likelihood Calibration (SLiC), can also be used to effectively learn from human preferences (\slichf).
Furthermore, we demonstrate this can be done with human feedback data collected for a different model, similar to off-policy, offline RL data.
Automatic and human evaluation experiments on the TL;DR summarization task show that \slichf significantly improves supervised fine-tuning (\sft) baselines.
Furthermore, \slichf presents a competitive alternative to the PPO RLHF implementation used in past work while being much simpler to implement, easier to tune and more computationally efficient in practice.
\end{abstract}

\section{Introduction}
While massively scaling model parameters and training compute of Transformer-based language models have led to impressive few-shot in-context learning \citep{gpt3,chowdhery2022palm},
reinforcement learning from human feedback fine-tuning (RLHF) can significantly improve generation quality as judged by humans.
This has been observed at all model scales for various downstream language generation tasks, such as abstractive summarization, dialogue, and creative writing \citep{openai_sum_hf,bai2022training,sparrow,instructgpt}.

For summarization in particular, multiple studies have shown that summaries generated by models tuned with \rlhf are preferred over the reference summaries in commonly used datasets \citep{openai_sum_hf,helm,goyal2022news}.
Reference summaries are often mined from web documents and might not have the highest quality or preferred style.
As a result, pure supervised learning, i.e. maximizing the likelihood of reference summaries given documents, is limited by the quality of reference summaries, thus additional feedback can improve models beyond the references. 
Commonly used reference-based metrics, such as ROUGE \citep{rouge}, only measure similarity between model generated and reference texts.
These reference-based metrics cannot measure quality improvement beyond the reference summaries.

To implement \rlhf, a reward model, $r_\phi(\seqx, \seqy)$, is trained on human preference data, $(\seqx, \seqy_0, \seqy_1, i) \sim D_{HF}$, collected via side-by-side human evaluation, where raters are asked to judge which of the two summaries $\seqy_0$ and $\seqy_1$ is better for document $\seqx$, i.e. $i\in\{0,1\}$.
If we denote the preferred summary as $\seqy^+$ and the other $\seqy^-$, the human feedback becomes $(\seqx, \seqy^+, \seqy^-)\sim D_{HF}$.
One common option for the training loss of the reward model used by \rlhf is:
\begin{equation}
    loss(r_\phi) = -\mathbb{E}_{(\seqx, \seqy^+, \seqy^-) \sim D_{HF}} [\log(\sigma(r_\phi(\seqx, \seqy^+) - r_\phi(\seqx, \seqy^-))]
\end{equation}
Reinforcement learning algorithms such as PPO \citep{ppo} are then used to refine a supervised fine-tuned model (\sft) to maximize the expected reward assigned by the reward model $r_\phi(\seqx, \seqy)$ \citep{ziegler2020finetuning,openai_sum_hf}.
A KL-penalty term is typically added to the loss to prevent the \rlhf model from diverging too far from the original supervised policy.

However, algorithms such as \rlhfppo introduce significant complexity to the training process by adding separate value and reward networks that may be comparable in size to the policy network. 
They are typically kept in memory to maximize training speed, which for a given memory budget significantly reduces the maximum size of trainable model.
Furthermore the optimization steps are significantly slower due to the use of roll-outs in the training loop, which involves sampling/decoding from the model. Hyper-parameter tuning and co-coordinating the PPO process is also more complex, requiring niche expertise.

Recently, another class of sequence-level contrastive methods \citep{brio, slic} seek to align model likelihood with an arbitrary, possibly non-differentiable reward, presenting an alternative to RL for optimizing the expected reward of samples.
\citet{slic} proposed Sequence Likelihood Calibration (\slic) to align a language model's sequence likelihood, $p_\theta(y | x)$, over decoded sequences according to their similarity to reference sequences.
The ranking \emph{calibration} loss contrasts a positive sequence $\seqy^+$  and a negative sequence $\seqy^-$, 
encouraging the model $P_\theta$ to assign more probability mass to positive compared to negative sequences:
\begin{equation}
  \lcal (\theta)  =  \max(0, \beta - \log P_\bt(\seqy^+ | \seqx) + \log P_\bt(\seqy^- | \seqx))  
\end{equation}
While the original \slic work used similarity to references as criteria for ranking, e.g. ROUGE \citep{rouge} and model embedding distances,
it can be replaced by an arbitrary, reference-less ranking function, $R(\seqy_0, \seqy_1, \seqx) \rightarrow \{0, 1\}$.
Particularly in this work, we use human preference as the ranking function, either by using off-policy preference data $D$ directly, or by training a predictive ranking model $R_\phi(\seqy_0, \seqy_1, \seqx)$ from $D$.

We call using \slic with this human preference ranking function \slichf and apply it using the human feedback data collected in \citet{openai_sum_hf}.
Our experiments show that \slichf also leads to improved summarization quality on the \reddit Summarization task as judged by humans, even though this feedback was collected for different models, similar to off-policy, offline RL.
While our T5-Large \citep{t5} (770M parameter) SFT model performs similarly to \citet{openai_sum_hf}'s 6B decoder-only SFT model, we are able to improve our model with \slichf such that it performs at least as well as \citet{openai_sum_hf}'s 6B \rlhfppo model as judged by humans.
Furthermore, applying \slichf to the T5-XXL 11B parameter \sft model \citep{t5} significantly improves results.

The primary contributions of this paper are showing:
\begin{itemize}
    \item how to apply \slic to learn from human preferences (\slichf), a simpler, more efficient yet competitive alternative to RLHF
    \item feedback/preference data from another model (off-policy) can be effectively leveraged by \slichf, making it unnecessary to collect costly new feedback data for our model
    \item providing a general \slichf recipe based on open-sourced T5 model that outperforms RLHF on the Reddit TL;DR summarization task

\end{itemize}

\section{Method}

In this work, we apply \slic \citep{slic} to improve a \sft model using human preference data 
$(\seqx, \seqy^+, \seqy^-)\sim D_{HF}$ in addition to the standard supervised fine-tuning data $(\seqx, \seqyt)\sim D_{SFT}$.

\subsection{Sequence Likelihood Calibration}
\label{sec:method_slic}

Following \citet{slic}, we first fine-tune a supervised model, $\supervisedmodel$, on $(\seqx, \seqyt)\sim D_{SFT}$, and then align the \sft model's sequence likelihood using the \slic approach which optimizes the following loss:
\begin{equation}
\mathcal{L}(\bt) = \sum \lcal(\bt,  \seqx, \seqyt, \mcand) +\lambda \lreg (\bt, \bt_{ft}; \seqx, \seqyt)
\end{equation}
where $\bt$ and $\bt_{ft}$ are the current and fixed \sft model weights, $\lcal$ and $\lreg$ are the calibration and regularization losses and $\{\seqyg\}_m$ are $\ndec$ sampled candidates from the SFT model.
More specifically, we choose the rank calibration loss and cross-entropy regularization loss for their simplicity and natural fit to pairwise human feedback data.
Thus the loss function of \slichf becomes the following:
\begin{equation}
\mathcal{L}(\bt) = max(0, \delta - \log P_\bt(\seqy^+ | \seqx) + \log P_\bt(\seqy^- | \seqx))  - \lambda \log P_\bt(\seqyt | \seqx)
\end{equation}
The first term is the calibration loss where $\seqx$ is the input sequence, $\seqy^+$ and $\seqy^-$ are positive and negative sequences, and $\delta$ is a hyper-parameter for the margin of the ranking loss. 
The second term is the cross-entropy loss, where $\seqyt$ is some target sequence and $\lambda$ is the regularization weight.
Cross-entropy loss encourages the model to stay close to the SFT model, similar to the KL term in used in \citet{openai_sum_hf}, however it does not need an extra copy of SFT weights.
The KL regularization term was also explored
in \citet{slic} but found to perform similarly. 
The choices of $\seqy^+$ and $\seqy^-$ are discussed in subsections \ref{sec:method_decode_rank} and \ref{sec:method_direct}. 
The choices of regularization target $\seqyt$ is discussed in \autoref{sec:method_regularization}. 

\subsection{\slichf with Sample and Rank}
\label{sec:method_decode_rank}

\citet{slic} samples candidates $\mcand \sim  \supervisedmodel$ from $D_{SFT}$'s training split, from which (positive, negative) pairs are determined.
We call this approach \emph{\slicdecoderank}.
To determine the rank, we consider two text-to-text models trained from the human preference data $D_{HF}$:

\paragraph{Trained Pointwise Reward model:}
Similar to \citet{askell2021general}, we binarize each ranked pair into a positive and a negative sequence, as shown in \autoref{fig:io}.
When training the reward model, input sequences are formatted as `[Context] ... [Summary] ... ' and target sequences are either `Good' or `Bad'.
At inference time, we compute the probability of token `Good' on the decoder side to score each of the $\ndec$ candidates in a list, and sample $\ndec$ positive/negative pairs from them.

\paragraph{Trained Pairwise Ranking model:} 
As shown in \autoref{fig:io}, we formulate the human feedback into a pairwise ranking problem with text-to-text format.
When training the ranking model, input sequences are formatted as `[Context] ... [Summary A] ... [Summary B]' and target sequences are among `A' or `B'.
At inference time, we use a tournament-style procedure to rank candidates in a list.
For example, given a list of 4 candidates $c_1, c_2, c_3, c_4$, we first rank $c_1, c_2$ and $c_3, c_4$ and then rank $winner(c_1, c_2), winner(c_3, c_4)$.
Given $\ndec$ candidates, the ranking model is called $\ndec-1$ times and $\ndec-1$ positive/negative pairs are yielded.

\input{figures/io_demo}

\subsection{\slichf Directly On Human Feedback}
\label{sec:method_direct}

We also consider a straight-forward approach of directly calibrating on positive and negative sequences from the human feedback dataset, $D_{HF}$, without a ranking or reward model. 
We call this approach \emph{\slicdirect}.
The obvious advantage of this approach is increased simplicity and efficiency from not training or using a ranking/reward model.
\slicdirect does not incur additional engineering costs in decoding from the \sft model and training a model to label the decodes. 
The drawback is that the off-policy human feedback data distribution might differ much from the \sft model's decode distribution.

\subsection{Regularization Term for Calibration}
\label{sec:method_regularization}

We consider two choices of target sequence $\seqyt$ for cross-entropy regularization.
The first choice is using $\seqyt$ in $D_{SFT}$ as regularization target.
The second choice is using the best ranked candidate from $\mcand$ as the regularization target.
Best ranked candidates can be selected using either the ranking model or the reward model.

\section{Experimental Results}

\subsection{Datasets}

We study \slichf on \reddit summarization datasets from \citet{openai_sum_hf}.
The dataset contains both fine-tune data $D_{SFT}$, human feedback data $D_{HF}$, along with their \sft and \rlhf model decodes which we use for comparison with our models.
$D_{SFT}$ is a filtered version of Reddit TL;DR dataset \citep{volske-etal-2017-tl}.
It contains 117k/6k/6k examples in train, validation and test splits.
$D_{HF}$ consists of 64k human preferences on decodes from multiple models.

\subsection{Experimental Hyper-parameters}

We conduct all experiments using T5 models \citep{t5} in the T5x framework \citep{t5x}.
In our ablation study, we choose a T5-large model (770M) as the generation model and T5-XXL (11B) as the ranking model and the reward model\footnote{We find that smaller T5 ranking/reward models do not converge reliably in our setup.}.
We train all generation models with batch size of 32 and ranking/reward models with batch size of 128.
Both are trained with default learning rate of $10^{-3}$.

We train the ranking model and the reward model on $D_{HF}$ training split, and picked checkpoints that have the highest accuracy on $D_{HF}$ validation split.
We fine-tune T5 models on $D_{SFT}$ training split, and pick checkpoints that have the lowest perplexity on $D_{SFT}$ validation split.

In calibration, we use learning rate of $10^{-5}$ and ranking margin $\beta$ of $1.0$. 
When calibrating models on their own decodes with \slicdecoderank, we sample 8 decodes with temperature of 0.7 and topk of 40 from fine-tuned only generation models. 

When evaluating our models, we use beam-search with beam size 4.
For automatic evaluation, we calculate the model decodes' win rate against human references measured by the T5-XXL ranking model on $D_{SFT}$ validation dataset.
Win rate is defined as the percentage of model decoded summaries preferred by the ranking model compared to human references.

\subsection{Reward Model and Ranking Model Accuracy}
\label{sec:ranker_accuracy}

Human feedback and human evaluation are done by raters comparing two summaries as it is
more reliable than pointwise rating.
We hypothesize that ranking model has an advantage over reward model because of its pairwise nature which aligns better with the task.
We train and compare a T5-XXL ranking model and a T5-XXL reward model (\autoref{sec:method_decode_rank}).
Results shows that our ranking model has accuracy of 73.23\% on $D_{HF}$ validation, about 2\% higher than our reward model which has accuracy of 71.34\%\footnote{Our ranking and reward models' accuracy are similar to the 6B reward model in \citet{openai_sum_hf}}. 

\subsection{\slic Ablation}

We conduct a set of experiments to ablate \slichf settings against baselines.
We use the ranking model as the main metric because of its higher correlation with human preferences demonstrated in \citet{openai_sum_hf}.
Selected settings are later verified with our human evaluation experiments in \autoref{sec:human_eval}.
We report ROUGE numbers just for reference purpose and do not use them to select models.
It is expected to see a drop in ROUGE numbers when learning from human feedback because it has less incentive to be similar to the reference texts.
Similar to \rlhf in \citet{openai_sum_hf}, we also observe an increase in average length of models and conduct a length controlled study in \autoref{sec:human_eval}.

\input{tables/ablation}

\subsubsection{\slichf vs Continue Fine-tuning on Filtered Data}

A simple way to learn from human feedback data is to convert it into \sft dataset and continue fine-tuning on it.
In general, we use the filtering approach which has similar performance to controlled generation approaches \citep{aharoni2022mface} but is cleaner to implement.
We consider three approaches to filter data for continued fine-tuning:

\begin{itemize}

\item keep only positive human feedback sequences and discard negative ones. 

\item decode 8 summaries from the \sft model, use the ranking model to select the best 1 out of 8 summaries by a tournament-style ranking approach.

\item decode 8 summaries from the \sft model, use the reward model to select the best 1 out of 8 summaries by scoring each and taking the one with the max score. 
\end{itemize}

As shown in \autoref{tab:ablation}, on \reddit dataset, continue fine-tune on positive human feedback data improves model win rate against reference slightly from 44.96\% to 51.65\%. 
In this experiment, we choose to use all human feedback without filtering for better models because this mimics a real world scenario where we have access to some human feedback data without the explicit knowledge of its quality. 
Continuing fine-tuning on best 1 out of 8 further improves win rate against reference to 60\%+ and using pairwise ranking model is slightly better than pointwise reward model for filtering.

\subsubsection{Apply \slichf Directly On Human Feedback Data}

With \slicdirect, we observed that even though calibration loss decreases as expected, sequence length keeps increasing and does not converge to a stable value.
On the other hand, \slicdecoderank robustly converges.
We hypothesize that \slicdirect is prune to out-of-distribution decodes generated by other models in the human feedback data.

When using the ranking model to select for the best checkpoint for \slicdirect, it has moderate length increment and has 82.92\% win rate against reference which is close to \slicdecoderank.
The engineering complexity of \slicdirect is almost the same as fine-tuning a model.
Therefore, it is a good candidate for quick experimentation on human feedback.

\subsubsection{Apply \slichf on Ranked Model Decodes}

As shown in \autoref{tab:ablation}, \slicdecoderank using the ranking model have about 3\% gain in win rate against reference compared to \slicdecoderank using the reward model.
This results aligns with the observation in \autoref{sec:ranker_accuracy} that the ranking model has higher agreement to human preference than the reward model.

For \slicdecoderank using the ranking or the reward model, using \sft targets or best ranked decodes as regularization doesn't show much difference.
This shows that \slicdecoderank is applicable even when there is no ground truth reference available.
The gain from continue fine-tuning on best ranked decodes in \autoref{tab:ablation} is not additive to \slichf.

\subsection{Human Evaluation}
\label{sec:human_eval}

We conduct side-by-side human evaluation between multiple systems using crowd-sourcing.\footnote{We use Amazon Mechanical Turk to set up the task and hire the raters} Given a document and 2-4 summaries, raters are tasked to assign a pointwise overall quality to each summary, select if the summary is factual or not, and choose the best summary.

Each task is replicated and judged by 3 different raters. To eliminate bias, we anonymize all the models and randomly shuffle order of summaries for each task.
We aggregate pointwise metrics by averaging the ratings across all 3 crowd workers, and we aggregate the choice metric using majority vote.

The human evaluation template and the rating instructions can be found in \autoref{appendix:human_eval}.

\subsubsection{\slichf Ablation Study}

We conduct a 4-way side-by-side human evaluation to confirm the ablation results in \autoref{tab:ablation}.
100 examples from the validation set are sampled from reference, \sft model, continue fine-tuning model and \slichf model (\slicdecoderank, using ranking model, regularized on best decodes).
As shown in \autoref{tab:human_eval_ablation}, \slichf is chosen as the best model 73\% of the time, has significantly higher average quality, and is the most factual model. In general, the average quality aligns well with the ranker win-rate from \autoref{tab:ablation}.

\autoref{fig:human_eval_len_controlled} shows the lengths controlled quality of \sft, continue fine-tuning and \slichf models, which clearly shows \slichf is preferred.
Length controlled quality study is similar to studies conducted in \citet{openai_sum_hf}, where mean scores are calculated among examples bucketed by their relative length to the reference.

\input{tables/human_eval_ablation}

\input{figures/human_eval_slic_ablation}

\subsubsection{\slichf vs \rlhfppo}

Correctly implementing and tuning the right hyper-parameters for the \rlhfppo algorithms in \citet{openai_sum_hf} are non-trivial tasks.
Instead of re-implementing the algorithms in our framework, we directly compare with the model decodes from \citet{openai_sum_hf}.

We first benchmark our T5-large \sft model against their 6B decoder-only \sft model in a two-way side-by-side human evaluation.
As shown in \autoref{fig:human_eval_slic_rlhf}, our \sft has slightly higher quality and win rate but it is not statistically significant.

Next we benchmark two variants of our T5-large \slicdecoderank models against the decoder-only 6B \rlhfppo model from \citet{openai_sum_hf}.
\slicdecoderank with reward model has similar performance as \rlhfppo and \slicdecoderank with ranking model is better than the \rlhfppo.
The summaries from \slichf models are slightly longer than the \rlhfppo model, and their length controlled win rate is similar to \rlhfppo as shown in \autoref{fig:human_eval_slic_rlhf}.

\input{tables/human_eval_slic_rlhf}

\input{figures/human_eval_slic_rlhf}

\subsection{Scaling Up \slic}

\input{tables/size}

We study 2 ways of scaling up the \slicdecoderank: (1) scaling up generation model parameters, (2) scaling up number of decoded candidates $\ndec$.
As shown in \autoref{tab:size}, scaling up generation model from 770M to 11B significantly improves both the \sft model and the \slichf model.
On the other hand, scaling up $\ndec$ from 8 to 64 does not help much.

\section{Further discussion on \slichf vs. \rlhfppo}

\subsection{Compute/Memory Efficiency and Parallelism}
\label{sec:efffiency}
We summarize the compute and memory efficiency differences between \slichf and \rlhfppo in \autoref{table:efficiency}.

\input{tables/efficiency}

In both \rlhfppo and \slicdecoderank we train an auxiliary ranking or reward model that is used to judge the quality of summaries.
However, \citet{openai_sum_hf} found that having separate policy and value networks worked significantly better, and thus contributes an extra auxiliary model, the same size as the reward model that is updated along-side policy updates. 

Furthermore, the policy, value, reward, and SFT models (all the same size in \citet{openai_sum_hf}) are used within the PPO training loop.
They are often held in hardware memory to ensure faster training steps.
Whereas in \slichf, the rewards can be computed completely in parallel and offline, thus using 1/4 the memory for model weights during training. Such memory savings could be re-purposed to train larger models.

\citet{openai_sum_hf} report using 1M episodes to conduct \rlhf training,
which corresponds to roughly the same number of decoded samples used in \slichf, ($\ndec=8$ per training example, 123,169 examples).
However, in practice \slichf decoding can be significantly faster because all the decoded samples use the same policy allowing for completely parallel decoding.
In contrast, with PPO the policy is updated every batch, limiting decoding parallelism to each batch (512, in \citep{openai_sum_hf}) as subsequent decoding is blocked on policy updates. Furthermore, PPO decoding occurs within the training loop leading to much longer optimization step times.
Whereas with \slichf step times are similar to fine-tuning, which is significantly faster as there is no decoding in the training loop.

Beyond the significant decoding parallelism gains, \slichf can make use of simple input encoding caching optimizations to reduce compute. 
Since the $\ndec$ decodes are sampled from the same \sft policy, the input sequence encoded states can be cached rather than recomputed.
In summarization and other tasks involving long contexts, this may be significant as input sequence length tends to be much longer than output.

\slichf has similar parallelism advantages in computing rewards per episode compared to \rlhf as the ranking can be computed outside the training loop instead of within.

\subsection{Pairwise Ranking vs Reward model}
RL algorithms seek to maximize the expected reward of trajectories, in this case the human judgement in quality of model summaries.
This reward function typically is assumed to be pointwise, whereas human preference data is collected pairwise to improve reliability.
Thus there is noise introduced in converting pairwise judgements into pointwise rewards, which can be estimated as the difference in ranking accuracy as in \autoref{sec:ranker_accuracy}.
Since \slichf only cares about the relative rank of two summaries, this pairwise-to-pointwise noise is avoided and we conjecture this helps \slichf (\autoref{tab:ablation}, \autoref{fig:human_eval_slic_rlhf}).

\subsection{The Value of States and Actions in Language}
For many tasks tackled using RL, rewards may be collected at the end of a trajectory (as in many Atari games) and the attribution of final reward to specific actions may be very important in learning to solve a task.
Typically when RL is applied to language as in the RLHF literature, the state is the prefix of the current text and the actions correspond to choosing the next token.
The value function's role is to estimate the goodness of a trajectory (e.g. summary) from a prefix/input, which is intuitively a very difficult task for human raters, and thus RL may also suffer from value function estimation noise.
In contrast, \slichf does not rely on such a sub-model and only uses the cleaner preference signal to drive parameter updates and leading to what we conjecture is more stable optimization.

\section{Related work}
RL has been used to optimize arbitrary reward in language generation such as BLEU for translation \citep{wu2016google} and ROUGE for summarization \citep{paulus2017deep}; however, while those metrics improved, human judgement of quality suffered due to metrics misalignment.

In an effort to better align the reward function with human judgement, many works used RL to align language models with a reward model trained to predict carefully collected human judgements \citep{ziegler2020finetuning, openai_sum_hf, instructgpt} using summarization as an initial proof-of-concept.
A KL penalty term, first used in \citet{jaques2017sequence}, is used as regularization to prevent the tuned model from departing from the initial supervised model, and is also used in \slic \citep{slic}. 

\citet{brio} propose BRIO, which has a similar intent as \slic \citep{slic} of rank-ordering model-generated decodes according to a reward function.
BRIO trains models to align length normalized sequence probability of generated decodes to their similarity to reference as measured by ROUGE using a list-wise loss function.
In contrast, and similar to RLHF, \slichf adapts the technique to align with a model trained to predict human preference given two summaries instead of their similarity to the reference.

\citet{bai2022constitutional} substitutes human preference data with judgements from a large language model, and calls it AI feedback (AIF).
SLIC-HF can also be used with AIF exactly in  the same way and is indifferent about the AI or human origin of the feedback.

\section{Conclusion}

In this work, we proposed \slichf that calibrates sequence likelihood on human feedback data.
Our experiments on the \reddit summarization task show that \slichf significantly improves supervised fine-tuning (\sft) baselines, and presents a competitive alternative to the \rlhfppo implementation of past work while being simpler to implement, easier to tune and computationally efficient.
Future work may include studying \slichf on other language generation tasks using other reward functions and/or non-human feedback.

\bibliography{anthology,custom}
\bibliographystyle{acl_natbib}

\newpage
\appendix

\input{appendix}

\end{document}

%% file: macros.tex
\newcommand{\slic}{SLiC\xspace}
\newcommand{\slichf}{SLiC-HF\xspace}
\newcommand{\sft}{SFT\xspace}
\newcommand{\rlhf}{RLHF\xspace}
\newcommand{\rlhfppo}{RLHF-PPO\xspace}
\newcommand{\slicdecoderank}{SLiC-HF-sample-rank\xspace}
\newcommand{\slicdirect}{SLiC-HF-direct\xspace}

\newcommand\reddit{Reddit TL;DR\xspace}

\newcommand\rouges{R1 / R2 / RL\xspace}

\newcommand\seqx{\mathbf{x}}
\newcommand\seqy{\mathbf{y}}
\newcommand\seqyt{\mathbf{y}_{\mathrm{ref}}}

\newcommand\seqyg{\hat{\mathbf{y}}}
\newcommand\mcand{\{\seqyg\}_m }

\newcommand\lcal{L^{\mathrm{cal}}}
\newcommand\lreg{L^{\mathrm{reg}}}
\newcommand\supervisedmodel{P_{\theta_{ft}}(\seqy | \seqx)}

\newcommand\bt{\mathbf{\theta}}

\newcommand\ndec{m}

%% file: figures/io_demo.tex
\begin{figure}[h!]
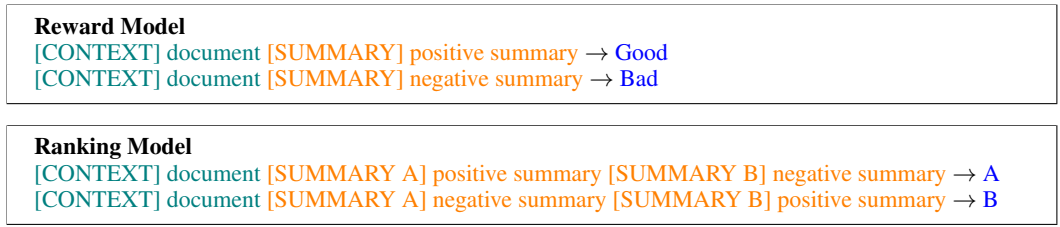

\centering


\begin{mdframed}
\small

\textbf{Reward Model}

\textcolor{teal}{[CONTEXT] document} 
\textcolor{orange}{[SUMMARY] positive summary}
$\rightarrow$
\textcolor{blue}{Good} 

\textcolor{teal}{[CONTEXT] document} 
\textcolor{orange}{[SUMMARY] negative summary}
$\rightarrow$
\textcolor{blue}{Bad} 
\end{mdframed}


\begin{mdframed}
\small

\textbf{Ranking Model}

\textcolor{teal}{[CONTEXT] document} 
\textcolor{orange}{[SUMMARY A] positive summary [SUMMARY B] negative summary}
$\rightarrow$
\textcolor{blue}{A}

\textcolor{teal}{[CONTEXT] document} 
\textcolor{orange}{[SUMMARY A] negative summary [SUMMARY B] positive summary}
$\rightarrow$
\textcolor{blue}{B}

\end{mdframed}


\caption{Training text-to-text reward model and ranking model.}
\label{fig:io}

\end{figure}

%% file: tables/ablation.tex
\begin{table}[tbh]
\caption{
Compare different methods to leverage human feedback data.
Ranker win rate is the T5-XXL ranking model's preference of choosing model decodes over reference texts.
}
\begin{center}

\small
\setlength{\tabcolsep}{4pt}
\begin{tabular}{lll ccc}
\hline
\multicolumn{3}{c}{Ablation} & \multicolumn{3}{c}{Metrics}  \\
method & human feedback form & regularization & {\# words} & \rouges & {ranker win rate} \\
\hline
\hline
\multicolumn{2}{l}{reference} & - & 27.11 & - & 50\% \\
\multicolumn{2}{l}{\sft} & - & 23.57 & 35.1/12.87/26.81 & 44.96\% \\
\hline

\multicolumn{6}{l}{continue \sft on filtered data} \\
 & positives from HF data  & - & 31.22 & 33.02/11.27/24.57 & 51.65\% \\
 & best decodes, by reward & - & 27.69 & 35.31/12.41/26.21 & 63.24\% \\
 & best decodes, by ranking & - &  28.26 & 35.39/12.69/26.56 & 65.43\% \\
\hline

\multicolumn{6}{l}{\slichf} \\
 & \slicdirect & \sft targets & 41.03 & 33.76/11.58/24.72 & 82.92\% \\
 & \slicdecoderank, by reward & \sft targets & 38.44 & 33.87/11.48/24.81 & 82.42\% \\
 & \slicdecoderank, by reward & best decodes & 38.58 & 34.07/11.59/24.92 & 83.52\% \\
 & \slicdecoderank, by ranking  & \sft targets & 37.96 & 34.49/11.92/25.35 & \textbf{86.21\%} \\
 & \slicdecoderank, by ranking & best decodes & 37.50 & 34.69/12.03/25.54 & \textbf{85.51\%} \\
\hline

\end{tabular}
\end{center}

\label{tab:ablation}
\end{table}

%% file: tables/human_eval_ablation.tex
\begin{table}[tbh]
\caption{4-way human evaluation to compare reference, \sft continue \sft on best decodes using ranking model, \slichf with pairs of decodes using ranking model.
}
\begin{center}

\small
\setlength{\tabcolsep}{4pt}
\begin{tabular}{lccccc}
\hline
 & reference & \sft & continue \sft & \slichf & same \\
\hline
chosen as preferred \% & 13\% & 5\% & 5\% & \textbf{73\%} & 4\% \\
average quality & 3.17 & 3.10 & 3.32 & \textbf{3.82} & - \\
is factual \% & 94.16\% & 94.85\% & 94.85\% & \textbf{96.56\%} & - \\
\hline

\end{tabular}
\end{center}

\label{tab:human_eval_ablation}
\end{table}

%% file: figures/human_eval_slic_ablation.tex
\begin{figure}[h]
\centering
\includegraphics[width=\textwidth]{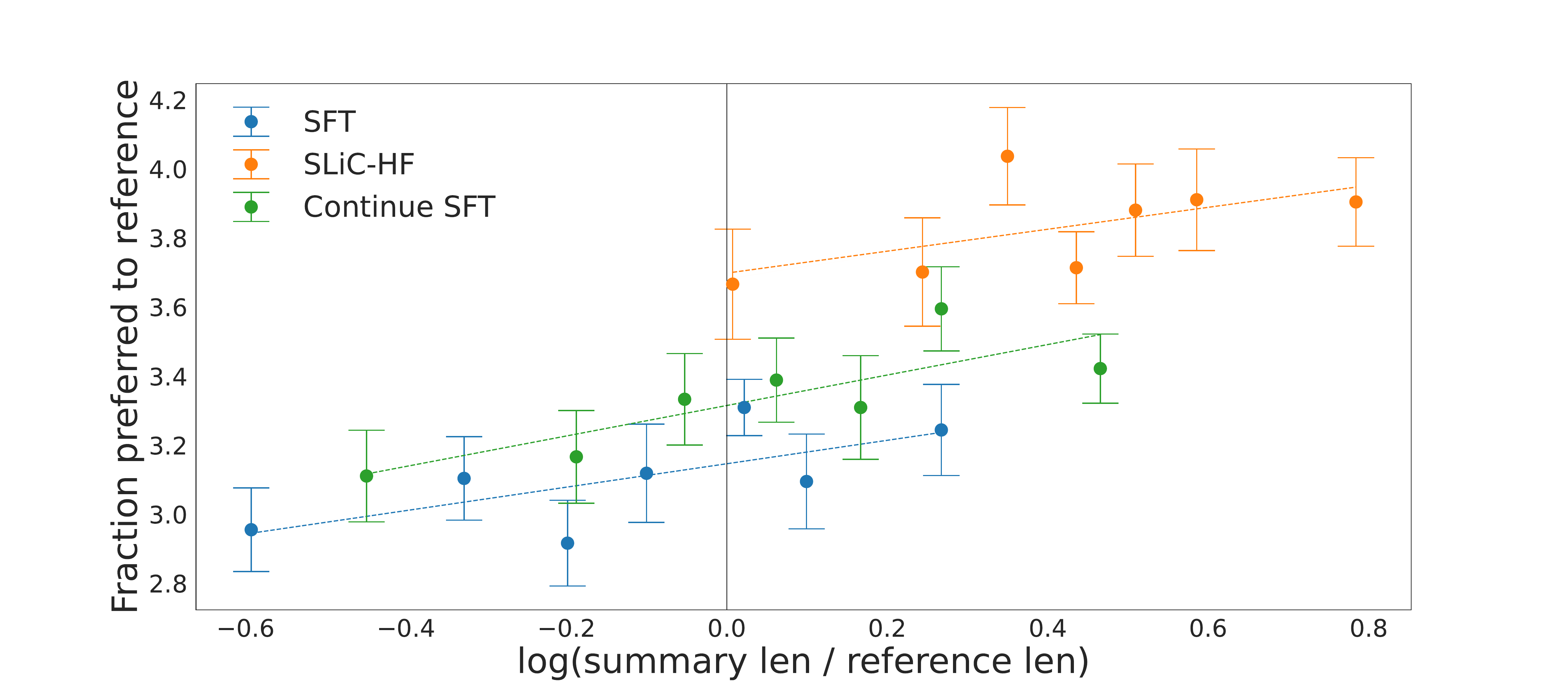}
\caption{Length bucketed average quality of \sft and \slichf against different baselines.}
\label{fig:human_eval_len_controlled}
\centering
\end{figure}

%% file: tables/human_eval_slic_rlhf.tex
\begin{table}[tbh]
\caption{
Three 2-way side-by-side human evaluations to compare our \sft baseline with \citep{openai_sum_hf}, and our \slichf models with the \rlhfppo model.
Statistically significant results are denoted with *.
}
\begin{center}

\small
\setlength{\tabcolsep}{4pt}
\begin{tabular}{cccc cc cc cc}
\hline
 \multicolumn{4}{c}{systems comparisons} & \multicolumn{4}{c}{human preference} \\
 \multicolumn{2}{c}{system A (ours)}  & \multicolumn{2}{c}{system B(\citep{openai_sum_hf})} & \multicolumn{2}{c}{win rate} & \multicolumn{2}{c}{quality} \\
 method & \# words &  & \# words & A & B & A & B \\
\hline
\sft (770M gen) & 23.7 & \sft (sup6B) & 24.6 & 56\% & 44\% & 3.59 & 3.48 \\
\slichf (700M gen, 11B ranking) & 36.9 & RLHF (sup6B\_rm6B) & 33.0 & 66\%* & 34\%* & 3.85* & 3.61* \\
\slichf (700M gen, 11B reward) & 38.4 & RLHF (sup6B\_rm6B) & 33.0 & 56\% & 44\% & 3.78 & 3.7 \\

\hline

\end{tabular}
\end{center}

\label{tab:human_eval_ablation}
\end{table}

%% file: figures/human_eval_slic_rlhf.tex
\begin{figure}[bth]
 \centering
 
 \begin{subfigure}
     \centering
     \includegraphics[trim={70 0 100 100}, width=0.48\textwidth]{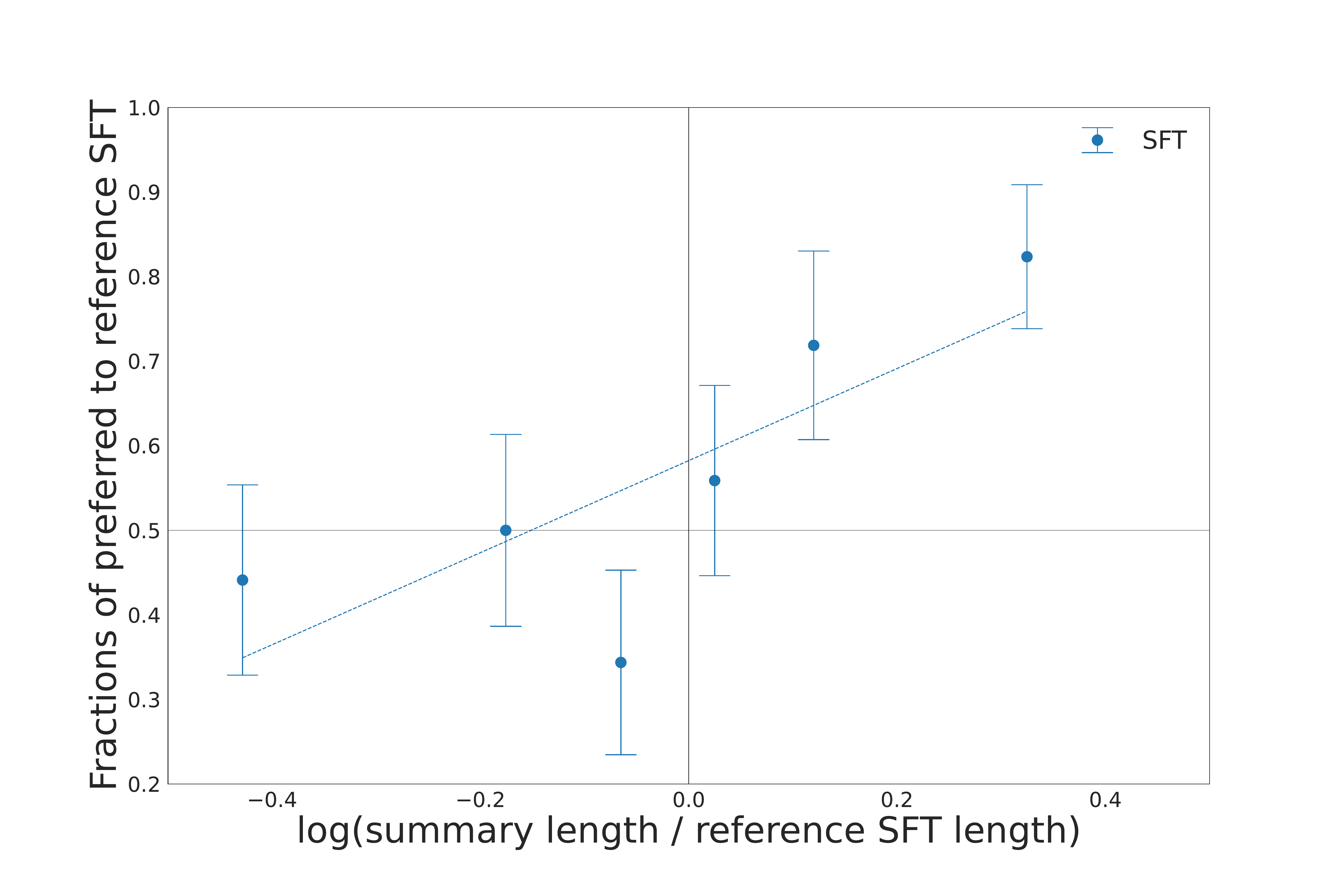}
 \end{subfigure}
 \hfill
 \begin{subfigure}
     \centering
     \includegraphics[trim={70 0 100 100}, width=0.48\textwidth]{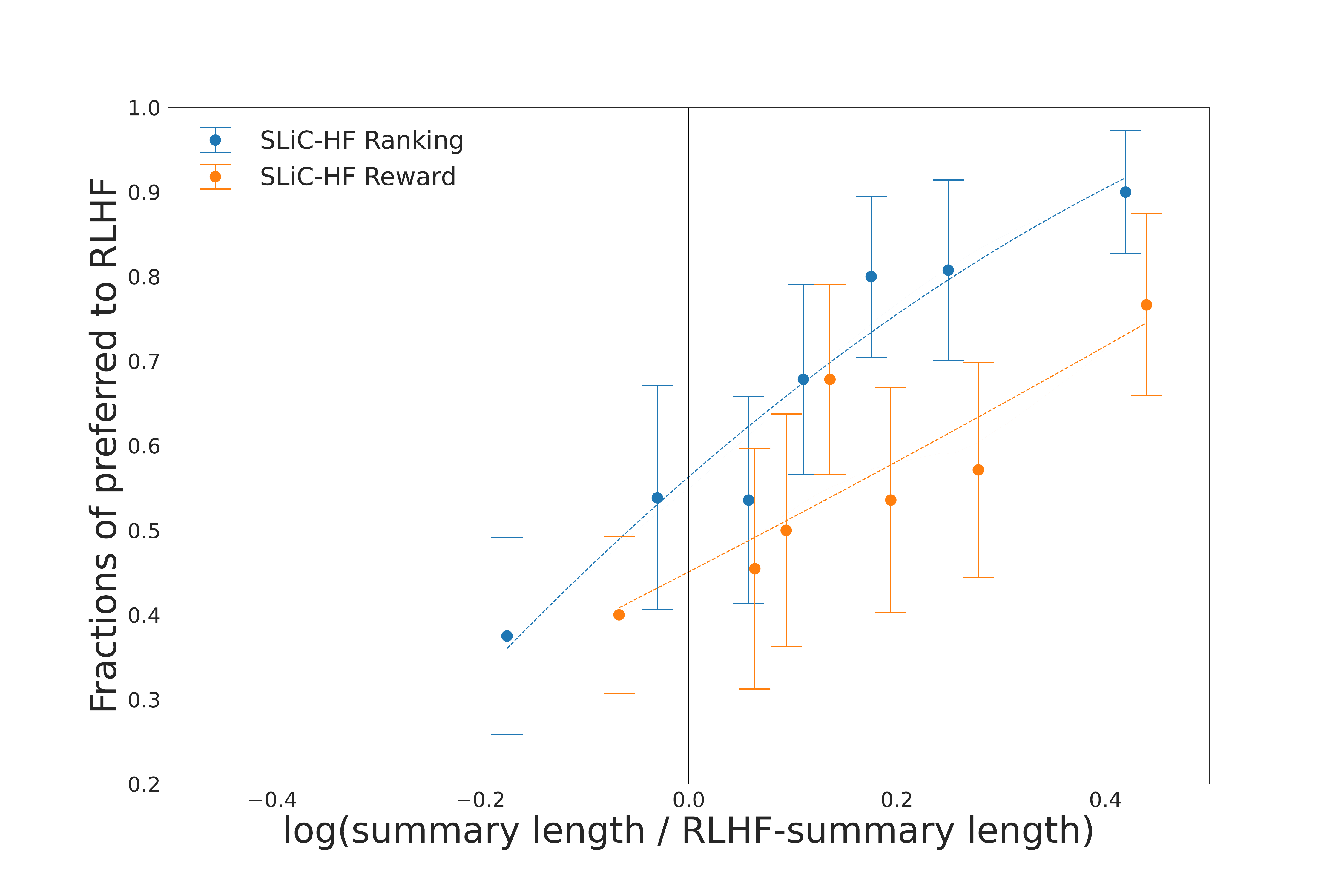}
 \end{subfigure}
 
\caption{Length bucketed average quality of \sft and \slichf against different baselines.}
\label{fig:human_eval_slic_rlhf}
\end{figure}

%% file: tables/size.tex
\begin{table}[tbh]

\caption{Effect of scaling up model parameters and number of candidates for \slicdecoderank.
}
\begin{center}

\small
\setlength{\tabcolsep}{4pt}
\begin{tabular}{lll ccc}
\hline
\multicolumn{3}{c}{Ablation} & \multicolumn{3}{c}{Metrics}  \\
method & \# params & $\ndec$ & {\# words} & \rouges & {ranker win rate} \\
\hline

\sft & 770M & 8 & 23.57 & 35.1/12.87/26.81 & 44.96\% \\
\sft & 11B & 8 & 24.07 & 36.45/14.11/28.38 & 62.34\% \\
\slichf & 770M & 8 & 37.96 & 34.49/11.92/25.35 & 86.21\% \\
\slichf & 770M & 64 & 40.53 & 34.14/11.70/25.11 & 86.41\% \\
\slichf & 11B & 8 & 36.90 & 35.83/12.87/26.63 & \textbf{96.10\%} \\
\hline

\end{tabular}
\end{center}

\label{tab:size}
\end{table}

%% file: tables/efficiency.tex
\begin{table}[h]
\centering
\caption{Compute and memory efficiency comparison. $p$ denotes the number of parameters in the policy network;}
\begin{tabular}{ccccc}
\hline

& \rlhfppo \citep{openai_sum_hf} & \multicolumn{2}{c}{\slichf} \\
&  & decode-rank & direct \\
\hline
Auxiliary models &  reward, value, \sft  &   ranking & -  \\ 
Decoded sequences & 1M  & ~800k & - \\
Parameter memory usage for training & $4p$ & $p$ & $p$ \\
Parameter updates per step & $2p$ & $p$ & $p$ \\
Parallel decoding & within batch & whole training set & - \\
Parallel reward & within batch & whole training set & - \\
Input encoding caching & no & yes & - \\
\hline
\end{tabular}
\label{table:efficiency}
\end{table}

%% file: appendix.tex
\section{Human Evaluation}
\label{appendix:human_eval}

\input{figures/mturk_screenshot}

See Figure~\ref{fig:mturk_example} for an example of the human evaluation task with 4 summaries. Summaries are randomly shuffled for each example and models are anonymized.

%% file: figures/mturk_screenshot.tex
\begin{figure}
    \centering
    \includegraphics[width=0.95\textwidth]{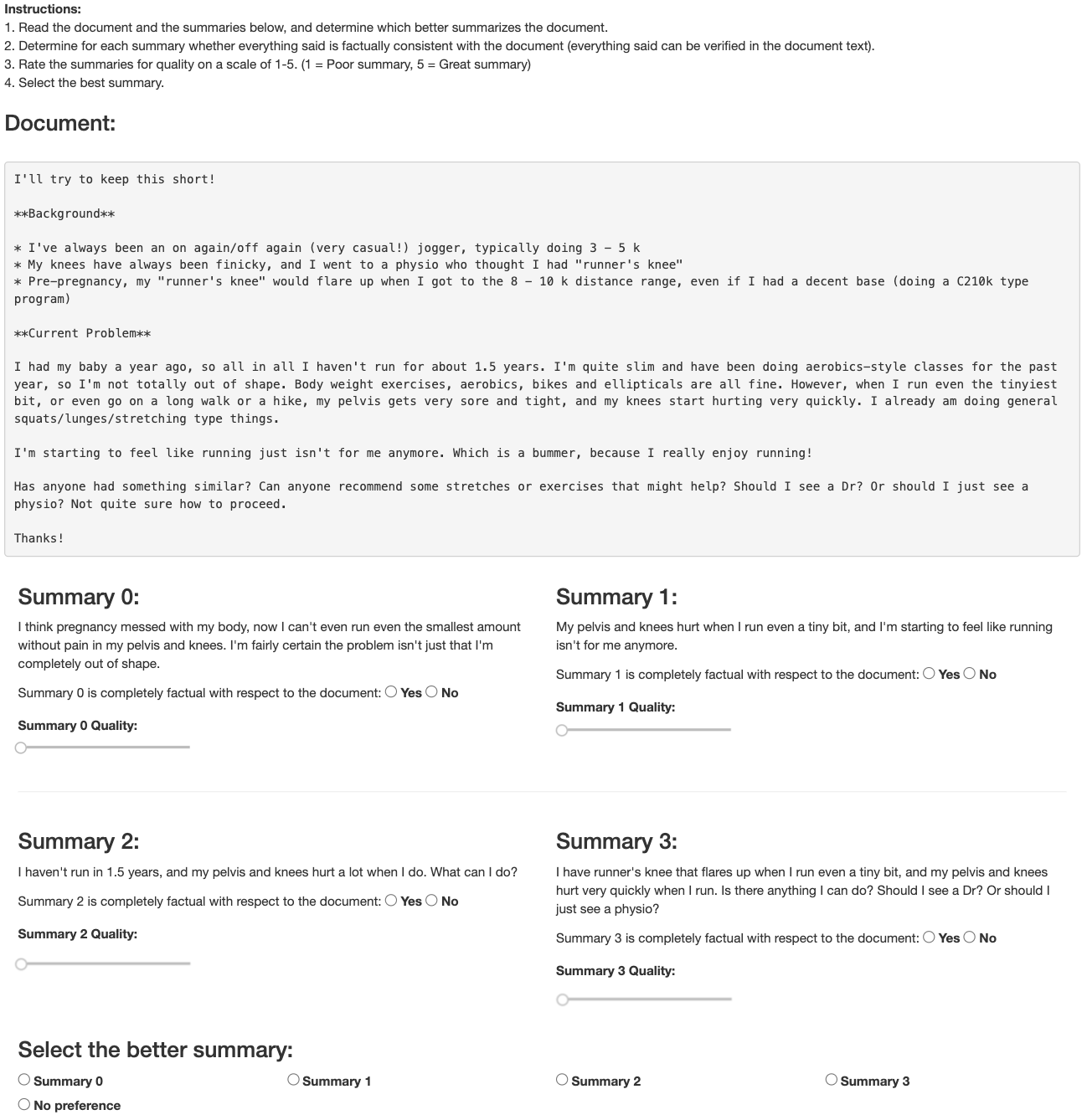}
    \caption{Example of human evaluation task.}
    \label{fig:mturk_example}
\end{figure}